\documentclass[runningheads]{llncs}
\PassOptionsToPackage{hyphens}{url}
\usepackage[hyphens]{url}
\usepackage{times}
\usepackage{latexsym}
\usepackage{graphicx}
\usepackage[T1]{fontenc}
\usepackage[utf8]{inputenc}

\usepackage{lscape}
\usepackage[dvipsnames]{xcolor}
\usepackage{color, colortbl, xcolor, soul}
\usepackage[dvipsnames]{xcolor}
\sethlcolor{WildStrawberry}

\usepackage{microtype}

\usepackage{todonotes}
\usepackage{graphicx}
\begin{document}

\title{A Comparative Study of Evaluation Metrics for Long-Document Financial Narrative Summarization with Transformers}

\author{
Nadhem Zmandar \inst{1}\orcidID{0000-0002-3087-6762} \and
Mo El-Haj \inst{1}\orcidID{0000-0002-6136-3898} \and
Paul Rayson \inst{1}\orcidID{0000-0002-1257-2191}
}

\authorrunning{N. Zmandar et al.}

\institute{
UCREL NLP group,\\
School of Computing and Communications,\\
Lancaster University, UK\\
\email{\{n.zmandar,m.el-haj,p.rayson\}@lancaster.ac.uk}
}
\maketitle              

\begin{abstract}
There are more than 2,000 listed companies on the UK's London Stock Exchange, divided into 11 sectors who are required to communicate their financial results at least twice in a single financial year. UK annual reports are very lengthy documents with around 80 pages on average.
In this study, we aim to benchmark a variety of summarisation methods on a set of different pre-trained transformers with different extraction techniques. 
In addition, we considered multiple evaluation metrics in order to investigate their differing behaviour and applicability
on a dataset from the Financial Narrative Summarisation (FNS 2020) shared task, which is composed of annual reports published by firms listed on the London Stock Exchange and their corresponding summaries. 
We hypothesise that 
some evaluation metrics do not reflect true summarisation ability and propose a novel BRUGEscore metric, as the harmonic mean of ROUGE-2 and BERTscore.
Finally, we perform a statistical significance test on our results to verify whether they are statistically robust, alongside an adversarial analysis task with three different corruption methods.

\end{abstract}

\section{Introduction}
With the proliferation of firms worldwide, the amount of financial disclosures and financial texts (or narratives) in various languages and formats has risen dramatically. Consequently, the study of natural language processing (NLP) methods that automatically summarize content has become a rapidly growing research area \cite{zmandar-etal-2021-financial} \cite{el-haj-etal-2022-financial}.\\
In fact, financial reporting and communication requirements have expanded significantly in recent years, particularly following the 2008 financial crisis. Financial communications and investor relations management are becoming increasingly critical to the financial markets and fund management industry. Regulated financial markets mandate that all listed companies regularly communicate their financial activities to stakeholders by publishing financial reports and other financial narratives.\\
Financial narratives are employed by firms to communicate with their stakeholders, including investors, shareholders, customers, employees, financial analysts, regulators, lenders, rating agencies, and suppliers. Through financial communications, stakeholders can assess how well the company is creating value.\\
The aim of this study is to create and evaluate summarization benchmarks for UK financial narratives, investigate the effect of long document methods, and examine their interactions with various metrics, including ROUGE, in order to assess their suitability for this domain. Additionally, we will introduce a statistical testing method for system-generated financial summaries and the novel BRUGEscore.

\section{Background}
Summarizing text is a complex task, and standard evaluation metrics such as accuracy, recall, and precision are not suitable for text summarization. In recent years, several metrics have been introduced that are specifically designed for evaluating the quality of machine-generated summaries. In this study, we used the following metrics:

\begin{itemize}
\item \textbf{ROUGE}: Recall-Oriented Understudy for Gisting Evaluation is a metric used to evaluate the quality of machine-generated summaries by comparing them with a set of human-produced reference summaries. ROUGE measures the number of overlapping textual units, such as n-grams or word sequences, between the generated summary and the reference summaries.

\item \textbf{BERTScore}: BERTScore is an embedding-based evaluation metric that aligns generated and reference summaries on a token level. Token alignments are computed to maximize the cosine similarity between contextualized token embeddings from the BERT transformer.

\item \textbf{BARTScore}: BARTScore is an unsupervised evaluation metric used for generative tasks such as machine translation, text summarization, and text generation. It offers a number of variants, depending on the language model used, that can be flexibly applied to evaluate generated text from different perspectives such as informativeness, fluency, or factuality.

\item \textbf{METEOR}: METEOR computes an alignment between candidate and reference sentences by mapping unigrams in the generated summary to 0 or 1 unigrams in the reference, based on stemming, synonyms, and paraphrastic matches.

\item \textbf{Bleurt}: Bleurt is a transfer learning-based metric for natural language generation that compares a candidate summary with a reference summary to determine how well the candidate summary conveys the meaning of the reference summary.

\item \textbf{BRUGEscore}: BRUGEscore is our novel proposed metric, calculated as the harmonic mean of ROUGE-2 and BERTscore. It combines elements of word overlap and embedding cosine similarity into a single score.
\end{itemize}

Table \ref{tab:metrics} provides a summary of the features of these metrics, including whether they are embedding-based or n-gram-based.

\begin{table}[h]
\centering
\fontsize{9}{9}
{\begin{tabular}{|l|c|l|l|}
\hline
\textbf{Metric} & \textbf{Embeddings} & \textbf{Language Model} & \textbf{n-gram} \\
\hline
ROUGE & No & N.A & n-gram\\
\hline
BERTScore & Yes & Roberta Large & 1-gram \\
\hline
BARTScore & Yes & Bart Large & 1-gram \\
\hline
METEOR & No & N.A & 1-gram \\
\hline
Bleurt & Yes & BERT-lg & Sequence \\
\hline
BRUGEscore & Yes & N.A & 2-gram \\
\hline
\end{tabular}}
\caption{Summary of the features of the evaluation metrics used in this study}
\label{tab:metrics}
\end{table}

\subsection{Related work}

Text summarization has shown promising applications in the financial domain \cite{el2019search}. Prior works in this field have explored a range of approaches. The Summariser system \cite{de2002financial} employed sentence linkage heuristics, while a query-based company-tailored summarization system was proposed in \cite{filippova2009company}. Recently, statistical features with heuristic approaches were used to summarise financial textual disclosures \cite{cardinaels2019automatic}. The Financial Narrative Summarisation (FNS) task of the Multiling 2019 workshop involved generating structured summaries from financial narrative disclosures. The FNS 2020 task \cite{elhaj-fns-2020} resulted in the first large scale experimental results and state-of-the-art summarisation methods applied to financial data, focusing on annual reports produced by UK firms listed on the London Stock Exchange (LSE).
The participating systems used a variety of techniques, ranging from rule-based extraction methods to traditional machine learning methods and high-performing deep learning models.

Prior works on UK annual report summarization include \cite{orzhenovskii-2021-t5}, who used a transformer-based encoder-decoder extractive summarisation approach based on the T5 pre-trained model. Abhishek Singh \cite{singh_point-5_2020} proposed a Pointer Network and T5-based summarization approach to extract relevant narrative sentences in a particular order to have a logical flow in the summary. Lei Li \cite{li_extractive_2020} used Determinantal Point Processes to build a Statistical learning Extractive Financial Narrative auto Summarizer. Jaime Baldeon Suarez, \cite{baldeon_suarez_combining_2020} combined financial word embeddings and knowledge-based features for financial text summarisation, and Moreno La Quatra \cite{la_quatra_end--end_2020} developed an end-to-end training framework for financial report summarisation in English. 


In comparison to prior works, we explore the impact of different transformer model architectures, the task and data used to pre-train transformer models, as well as correlations between automated metrics within the task of summarising UK annual reports. Our work is distinct as UK annual reports are long, unstructured in plain text, technically written, and subjective. Our study aims to address the challenging components of Financial Narrative Summarisation, and this effort is further promoted by the 2021 Financial Narrative Summarisation task (FNS 2021) in the FNP 2021 workshop.

To address the memory efficiency issue of transformers, we cannot simply pass the entire input annual report and gold standard to the model and fine-tune it. Instead, we need to determine which parts of the report to pass to the transformer. Through dataset analysis, we found that the gold standards are typically extracted from the first third of the report, where the chairman or CEO message and financial highlights are usually located. Therefore, we will pass the first k tokens to the model, where k depends on the model architecture, pre-training, and memory efficiency. Then, the model will be trained to predict the first n tokens of the system summary. On the test dataset of 500 UK annual reports, the model will predict the first n tokens, and we will continue the extraction of the remaining k tokens by determining which part of the report matches the predicted n tokens. This approach transforms the summarization problem into a task of predicting the start of the summary, allowing us to adapt sequence-to-sequence transformer models to summarize long documents where the reference summary is a continuous extracted part of the original text. We refer to this technique as the block-based summarization approach. This technique surpasses the memory efficiency issue of some transformers and is motivated by the fact that reference summaries are extracted from the financial annual report as a block. To our knowledge, this is the best approach for adapting encoder-decoder transformer models to summarize long documents.

We describe several techniques for summarization in this paper, including transformer-based \cite{orzhenovskii-2021-t5}, reinforcement learning-based \cite{zmandar-etal-2021-joint}, unsupervised learning using LSA, BERT extractive \cite{miller2019leveraging}, and SBERT extractive summarisation \cite{reimers-gurevych-2019-sentence}. We also compare the results of these techniques to four toplines and baseline summarizes, as we show later in the papers, and finally, we use Lead-1000 (the first 1000 words) as a strong baseline summarizer \cite{radev2004centroid}.


The block-based summarization approach is described as a method of adapting sequence-to-sequence transformer models to summarize long documents where the reference summary is a continuous extracted part of the original text \cite{orzhenovskii-2021-t5}. RL-based summarization is also discussed as a suitable approach for maximizing a predefined metric \cite{zmandar-etal-2021-joint}. Finally, we briefly explain LSA \cite{Gong2001GenericTS}, BERT extractive \cite{miller2019leveraging}, and SBERT extractive \cite{reimers-gurevych-2019-sentence} as unsupervised techniques that can be used to identify important sentences in a document.

\section{Dataset}
The dataset used for this study is composed of UK annual reports in English from the financial summarisation shared task (FNS 2021) \cite{zmandar-etal-2021-financial}. It contains 3,863 annual reports for firms listed on the London Stock Exchange (LSE) covering the period between 2002 and 2017, with an average length of 52,000 tokens. The dataset also includes 9,873 gold standard summaries. The dataset is randomly split into training (75\%), testing, and validation (25\%). Table \ref{tab:gold_standard} shows the dataset details.

\begin{table}[h]
\centering
{\begin{tabular}{|l|r|r|r|}
\hline
Data Type & Train & Validate & Test\\
\hline
Report full text & 3,000 & 363 & 500 \\
\hline
Gold summaries & 9,873 & 1,250 & 1,673\\
\hline
\end{tabular}}
\caption{FNS 2021 Shared Task Dataset}
\label{tab:gold_standard}
\end{table}

The dataset used for training presents several gold standard summaries for each annual report (between one and five) \cite{zmandar-etal-2021-financial}. We wanted to use multiple references to make the process more objective since we did not have a human-generated reference summary as a good gold standard. The gold standards used in this study were Financial Highlights, Letter to the Shareholders, Financial Statements, and Auditor’s Report. 

\section{Experimental Work}

In our experiments, we used various transformer models used in the study, including the T5 transformer \cite{raffel_exploring_2020}, LongFormer Encoder-Decoder \cite{beltagy_longformer_2020}, as well as BART, Pegasus, and BERT \cite{lewis-etal-2020-bart} \cite{zhang_pegasus_2020} \cite{devlin_bert_2019}.

In our study, we investigated whether using multiple gold standard summaries would improve the performance of summarization models. To fine-tune the models, we first considered the issue of gold summary standards. We trained T5, Pegasus, and BART using two different strategies. The first strategy involved using all available gold standards, which meant creating multiple pairs for each report. The second strategy was to choose only one gold summary that maximized the ROUGE metric \cite{zmandar-etal-2021-joint}, which was the aim of the FNS task. Our preliminary study found that training on a multi-referenced dataset did not significantly improve the ROUGE result and was computationally expensive. Therefore, we chose to train our models using only one reference summary per annual report. We set our reward function as ROUGE-2 and selected the gold standard summary that maximized the ROUGE-2 score with the annual report. This enables our system sumamrisers to maximize the Rouge metric with all the reference summaries.

For hyperparameter search, a comprehensive grid search is a common approach. However, due to the significant computational power and time required, we opted for a simpler strategy. We selected hyperparameters that maximize the input length and target length for our models, as detailed in Table \ref{tab:hyperparameters}.
\begin{table*}[ht]
  \centering
  {\begin{tabular}{|l|l|r|r|r|r|}
  \hline
    Transformer & model\_name & max\_input & max\_target & batch\_size & train\_epochs \\
    \hline
    T5 & t5-small & 4096 & 512  & 4 & 5\\
    \hline
    LED base & allenai/led-base-16384 & 8000 & 1000  & 4 & 5 \\
    \hline
    LED large & allenai/led-large-16384 & 4096 & 512  & 4 & 5 \\
    \hline
    Pegasus & google/pegasus-large  & 1024 & 256  & 4 & 5 \\
    BART & facebook/bart-base & 1024 & 128  & 4 & 5\\
    \hline
  \end{tabular}}
  \caption{description of hyperparameters during training on the FNS dataset}
  \label{tab:hyperparameters}
\end{table*}
In this study, we used metrics that support multiple references to evaluate the performance of our models. To compute the score between the system summary and all the gold standards, we used the Rouge.2.0 java jar\footnote{\url{https://github.com/kavgan/ROUGE-2.0}} file for ROUGE evaluation. We removed English stop-words but did not use an English stemmer. For other metrics, we used the implementation from the original authors or the implementation of the Hugging Face team on the datasets library. Table \ref{tab:summary_results} provides a summary of the results. We compared the best version of each transformer model with different baselines and toplines, as well as our new BRUGEscore. F1 scores were reported for each metric, including four variants of the rouge score (R1, R2, R-L, R-SU4), BERT and BART scores, Meteor and Bleurt scores. To compute the embedded representation, we used the Roberta-large-mnli and Bart-large-mnli language models for BERTScore and BARTScore, respectively.

The results suggest that model-based metrics give good results on the financial dataset, and that Bleurt is not a suitable metric to evaluate system performance. T5 is the best text-to-text model for the dataset, performing well alongside Longformer Encoder-Decoder. LED base is memory-efficient and performs very well on the dataset, while LED Large did not perform as well due to limited GPU memory. The BRUGEscore shows a harmonic mean between the Rouge2 score and BERT score, giving an equilibrium between sentence semantics and exact 2-gram matching. Lead-1000 is a strong benchmark in this task, indicating the superiority of transformer-based summarisation over deep learning and reinforcement learning methods.
\begin{figure}[ht]
    \centering
\includegraphics[width=\linewidth]{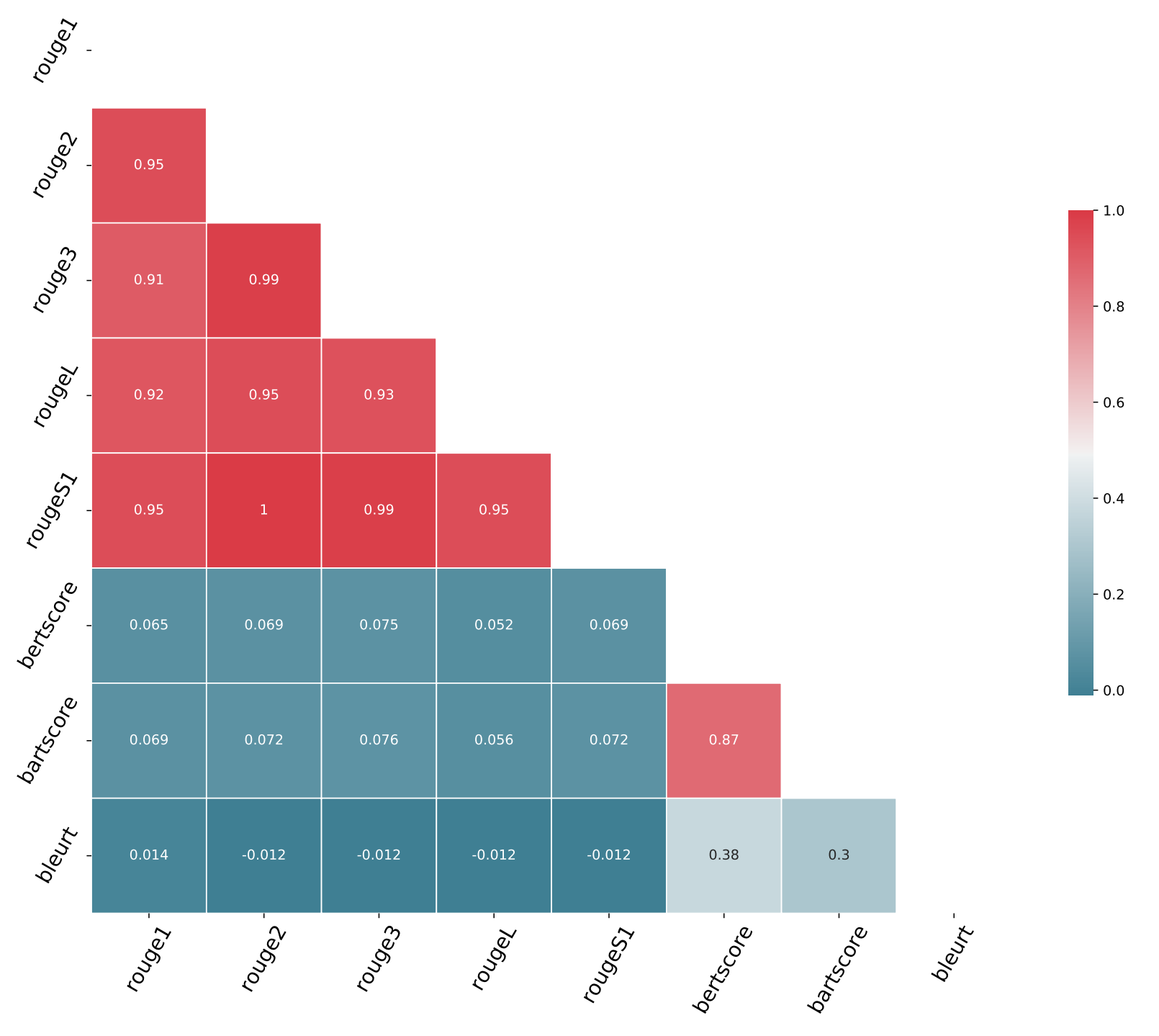}
    \caption{Correlation Matrix of Scores Produced using T5}
    \label{fig:corrplots}
\end{figure}

Figure \ref{fig:corrplots} shows the correlation matrix of different evaluation metrics' scores using summaries produced by the T5 transformer models which was pre-trained on the FNS test dataset\footnote{We only display the T5 matrix as it aligns with our conclusion, and the matrices of the other transformers exhibit similar patterns.}. The correlation plot shows that the different variants of the ROUGE metric are highly correlated, motivating the use of only one ROUGE variant in the evaluation process. Additionally, BERTScore and BARTScore are highly correlated, while BERTscore, BARTscore, and Bleurt are not correlated with the different variants of ROUGE.

\begin{table*}[ht]
\fontsize{8}{8}
\centering
\begin{tabular}{|l|c|c|c|c||c|c||c|c|c|}
\hline
\textbf{System / Metric} & \textbf{R-1/F} & \textbf{R-2/F} &  \textbf{R-L/F} & \textbf{R-SU4/ F} & \textbf{BE/F} & \textbf{BA/F} & \textbf{bleurt} & \textbf{meteor}  &  \textbf{BR}  \\
\hline

T5-Small-96 &   0.496 &   0.374 &   0.487 & 0.417 &   0.910 &   0.830 &    -0.836972 &   0.184 &  0.530\\

\hline
LED-base-128 &   0.492 &   0.370 &   0.484 &   0.413 &   0.899 &   0.816 &   -0.849750 &   0.182 & 0.524 \\

\hline
Pegasus &   0.476 &   0.350 &   0.467 &   0.394 &   0.847 &   0.759 &   -0.925372 &   0.174 & 0.495 \\
 
\hline
BART &  0.453 &   0.317 &   0.440 &   0.365 &   0.852 &   0.774 &   -0.928474 &   0.176 & 0.462 \\
\hline
\hline

\hline  
Lead-1000 &   0.443 &   0.307 &   0.431 &   0.356 &   0.774 &   0.694 &    -1.039358 &   0.162 & 0.440  \\

\hline
RNN-LSTM-RL &   0.459 &   0.270 &   0.431 &   0.268 &   0.761 &   0.647 &   -1.027724 &   0.175 & 0.399 \\

\hline
MUSE-topline &   0.433 &   0.234 &   0.419 &   0.253 &   0.756 &   0.655 &   -1.045138 &   0.163  & 0.357 \\

\hline
LSA &   0.321 &   0.140 &   0.287 &   0.187 &   0.782 &  0.651 &   -0.945594 &   0.160 & 0.237 \\

\hline
\hline

SBERT-extractive &   0.322 &   0.139 &   0.276 &   0.187 &   0.781 &   0.647 &   -0.973918 &   0.159 & 0.236 \\

\hline
BERT-extractive &   0.312 &   0.134 &   0.263 &   0.182 &   0.771 &   0.632 &   -0.987254 &   0.121 & 0.228 \\

\hline
LexRank &   0.264 &   0.120 &   0.253 &   0.140 &   0.732 &   0.580 &   -1.051438 &   0.088 & 0.206 \\

\hline
POLY-BASELINE &   0.274 &   0.105 &   0.212 &   0.135 &   0.723 &   0.565 &   -1.060618 &   0.109 & 0.183 \\

\hline
TextRank &   0.172 &   0.070 &   0.242 &   0.079 &   0.727 &   0.576 &   -1.074088 &   0.088 & 0.128\\

\hline
\end{tabular}
\caption{F-measure scores for Rouge-1, Rouge-2, Rouge-L, SU4, BERTScore, BARTScore, Bleurt, and Meteor, ranked based on Rouge-2 F1 measure. The abbreviations used are BE for BERT score (roberta-large-mnli), BA for BART score (bart-large-mnli), and BR for BRUGEscore.}
\label{tab:summary_results}
\end{table*}

\section{Statistical significance}

To compare the performance of two algorithms or models, we need to prove that the evaluation metric, denoted by `e', is greater for one system than the other. However, this is not sufficient as we also need to check the statistical significance of the difference in performance between the two algorithms. The common practice in NLP is to claim superiority of one algorithm over another only if the difference in results is statistically significant. To do that, we use significance levels and p-values to determine whether the test results are statistically significant, to avoid false discoveries. We follow the guidelines from the Hitchhiker’s Guide to Testing Statistical Significance in NLP" \cite{dror2018hitchhiker}. We model our problem as a ``no difference'' (null hypothesis H0) or ``difference'' (H1) and choose the bootstrap test to verify the significance of our results. We apply our test to the difference between the series of results generated by each system, report the p-values of ROUGE-2, ROUGE-L, BERTscore, and Bleurt score as shown in Tables \ref{tab:BE-pvalues} to \ref{tab:RL-pvalues} in Appendix A. We present the p-values of ROUGE-2, ROUGE-L, BERTscore, and Bleurt score in the tables obtained through the Bootstrap method. These p-values, when compared to the significance level (0,1), indicate the significance of the performance difference between the two systems. Cells that are not coloured red indicate a statistically insignificant difference, allowing us to claim with 90\% confidence that system one system outperforms the other using a specific metric.

\section{Adversarial analysis}
To assess the robustness of the metrics, we also conducted an adversarial analysis on the predicted summaries. Adversarial attacks are text perturbations designed to test the effectiveness of the metrics. Our experiments involved corrupting a set of summaries generated by the T5 small model, which was the best-performing model on the test dataset. We tested the ability of the metrics to resist different sources of noise using \textbf{a) BERT mask-filling, b) word-dropping, and c) word permutation} methods. BERT mask-filling and word-dropping are derived from the method used to pre-train BLEURT, while word permutation tests the metrics' sensitivity to syntax by swapping the ordering of two adjacent tokens in the summary. We chose four values of chunks to avoid creating a bias in the distribution of corrupted tokens: 4, 6, 8, and 10. By uniformly distributing the corruption across the text, we can evaluate how well the metrics reflect the difference between the corrupted and uncorrupted text. We anticipate that higher-quality summaries will be more robust to noise. Word-dropping simulates some of the common issues that can arise with extractive summarization. BERT mask-filling is a denoising encoding task that is challenging for BERT score since it assumes that the predicted word by a BERT model is better in this context than the original word in the system summary. Word permutation will penalize the n-gram based metrics but will favour model-based metrics like BERT score and BART score.

To compare the original and corrupted summaries, we use a strict comparison where the original summary must be strictly better than the corrupted one. Table \ref{tab:aversarial_10} shows the results for the three adversarial tasks with a chunk length of 10. The accuracy value represents the percentage of non-corrupted summaries that received better scores than their corrupted counterparts. An accuracy of 0.00 indicates that the corrupted and non-corrupted summaries received the same scores, as with ROUGE-1 during the word permutation corruption test. This is because ROUGE-1 is insensitive to syntax.

BERTScore and BARTScore achieved an accuracy score of 60\% across the three different tasks. These results suggest that \textbf{ROUGE} is better suited for extractive summarization while model-based metrics are more suitable for abstractive summarization. ROUGE evaluates summaries on a word-by-word basis, whereas model-based metrics consider the context as a whole. The results also show that ROUGE-2 performed best on the word permutation and BERT mask-filling tasks, while ROUGE-3 performed best on the word dropping task. When the corruption is applied to a single token in a sentence, it disrupts the n-gram sequence, which impacts ROUGE-n when n is greater than 1. Bleurt returned poor results, confirming that it is more suitable for comparing different models than evaluating a single model.

\begin{table*}[ht]
\fontsize{9}{8}
\centering
\begin{tabular}{|l|r|r|r|r|}\hline
\textbf{Metric} & \textbf{Word dropping\_10 (\%)} & \textbf{Word Permutation\_10 (\%)} & \textbf{Bert Mask filling\_10(\%)} \\
\hline
ROUGE-1 & 0.826 & 0.000 & 0.982  \\
\hline
ROUGE-2 & 0.958 & 1 & 0.99 \\
\hline
ROUGE-3 & 0.968 & 0.998 & 0.992 \\
\hline
ROUGE-S1 & 0.958  & 1 & 0.99 \\
\hline
ROUGE-S2 & 0.946 & 0.996 & 0.992  \\
\hline
ROUGE-L & 0.922 & 0.978 & 0.99 \\
\hline
ROUGE-SU4 & 0.88 & 0.994 & 0.992 \\
\hline
BERTScore & 0.608 &  0.63 & 0.668 \\
\hline
BARTScore & 0.636 &  0.6 & 0.656  \\
\hline
BLEURT & 0.556 & 0.632 & 0.574 \\
\hline
\end{tabular}
\caption{Mean accuracy by metric on the three corruption tasks. We apply three types of corruptions on the system generated summaries. We create a corruption every 10 chunks. Each metric is used to score the original and the corrupted versions of these summaries. This task should give the uncorrupted version a higher score to make sure that the metric is sensitive to corrupted summaries. The results reported shows the accuracy by metric on this task. All standard deviations were small (less than 0.2\%).
The experiments were performed on the FNS dataset using  the best performing system which is the small version of T5 transformer}
\label{tab:aversarial_10}
\end{table*}

\section{Conclusion and Future Work}
This paper tackled the task of automatic financial extractive summarisation of UK annual reports using various transformer models and unsupervised baselines. We proposed a set of model-based evaluation metrics, including a new metric called BRUGEscore, which outperformed ROUGE metric variants. We analyzed the results and performed adversarial analysis on the system-generated summaries to verify the robustness of the metrics. In the future, we plan to perform a human evaluation task on our dataset, measure the correlation with existing evaluation metrics, and work on improving the quality of the reference summaries. All PyTorch models are hosted on a private huggingface repository and will be released once the paper is accepted.

\section{Limitations}
The lack of gold standards, specifically human-generated summaries by domain experts, is the biggest technical challenge facing the financial text summarisation research community. We currently use extracted sections from annual reports as gold summaries. Furthermore, the results are limited to this English dataset, and the performance of evaluation metrics on other languages cannot be guaranteed, especially for language model-based models that are pretrained on English. Financial datasets are also large and scalable, requiring significant computational capacities. Finally, the jargon used in financial disclosures is different from 'general' language, and there is an urgent need to pre-train financial-specific language models for use in such studies.

\bibliographystyle{splncs04}
\bibliography{custom}

\section*{Appendix A}\label{sec:appendix}

\begin{table*}
\resizebox{\textwidth}{!}{

\begin{tabular}{lccccccccccccccccccccc}


T5-Small-96 &            &            &            &            &            &            &            &            &            &            &            &            &            &            &            &            &            &            &            &            &            \\

LED-BASE-128 &     0.0448 &            &            &            &            &            &            &            &            &            &            &            &            &            &            &            &            &            &            &            &            \\

LED-BASE-256 &     0.0161 &     \cellcolor{red}{0.3352} &            &            &            &            &            &            &            &            &            &            &            &            &            &            &            &            &            &            &            \\

LED-BASE-1000 &     0.0042 &     \cellcolor{red}{0.1930} &     \cellcolor{red}{0.3210} &            &            &            &            &            &            &            &            &            &            &            &            &            &            &            &            &            &            \\

BART &     0.0000 &     0.0000 &     0.0000 &     0.0000 &            &            &            &            &            &            &            &            &            &            &            &            &            &            &            &            &            \\

mBART &     0.0000 &     0.0000 &     0.0000 &     0.0000 &     \cellcolor{red}{0.4943} &            &            &            &            &            &            &            &            &            &            &            &            &            &            &            &            \\

PEGASUS &     0.0000 &     0.0000 &     0.0000 &     0.0000 &    \cellcolor{red}{0.2440} &     \cellcolor{red}{0.2429} &            &            &            &            &            &            &            &            &            &            &            &            &            &            &            \\

T5-MULTI-REFERENCES &     0.0000 &     0.0000 &     0.0000 &     0.0000 &     \cellcolor{red}{0.2373} &     \cellcolor{red}{0.2306} &     \cellcolor{red}{0.4825} &            &            &            &            &            &            &            &            &            &            &            &            &            &            \\

T5-Small-256 &     0.0000 &     0.0000 &     0.0000 &     0.0000 &     0.0106 &     0.0064 &     0.0378 &     0.0351 &            &            &            &            &            &            &            &            &            &            &            &            &            \\

T5-Small-512 &     0.0000 &     0.0000 &     0.0000 &     0.0000 &     0.0015 &     0.0022 &     0.0112 &     0.0059 &     \cellcolor{red}{0.2900} &            &            &            &            &            &            &            &            &            &            &            &            \\

PEGASUS-MULTI-REFERENCES &     0.0000 &     0.0000 &     0.0000 &     0.0000 &     0.0000 &     0.0000 &     0.0000 &     0.0000 &     0.0000 &     0.0001 &            &            &            &            &            &            &            &            &            &            &            \\

BART-MULTI-REFERENCES &     0.0000 &     0.0000 &     0.0000 &     0.0000 &     0.0000 &     0.0000 &     0.0000 &     0.0000 &     0.0000 &     0.0000 &     0.0077 &            &            &            &            &            &            &            &            &            &            \\

LSA &     0.0000 &     0.0000 &     0.0000 &     0.0000 &     0.0000 &     0.0000 &     0.0000 &     0.0000 &     0.0000 &     0.0000 &     0.0550 &     0.0000 &            &            &            &            &            &            &            &            &            \\

SBERT-EXTRACTIVE &     0.0000 &     0.0000 &     0.0000 &     0.0000 &     0.0000 &     0.0000 &     0.0000 &     0.0000 &     0.0000 &     0.0000 &     0.0300 &     0.0000 &     \cellcolor{red}{0.2948} &            &            &            &            &            &            &            &            \\

LEAD-1000 &     0.0000 &     0.0000 &     0.0000 &     0.0000 &     0.0000 &     0.0000 &     0.0000 &     0.0000 &     0.0000 &     0.0000 &     0.0000 &     0.0000 &     0.0093 &     0.0172 &            &            &            &            &            &            &            \\

LED-LARGE &     0.0000 &     0.0000 &     0.0000 &     0.0000 &     0.0000 &     0.0000 &     0.0000 &     0.0000 &     0.0000 &     0.0000 &     0.0000 &     0.0000 &     0.0068 &     0.0132 &     0.4912 &            &            &            &            &            &            \\

BERT-EXTRACTIVE &     0.0000 &     0.0000 &     0.0000 &     0.0000 &     0.0000 &     0.0000 &     0.0000 &     0.0000 &     0.0000 &     0.0000 &     0.0000 &     0.0000 &     0.0000 &     0.0000 &     \cellcolor{red}{0.1234} &     \cellcolor{red}{0.1335} &            &            &            &            &            \\

RNN-LSTM-RL &     0.0000 &     0.0000 &     0.0000 &     0.0000 &     0.0000 &     0.0000 &     0.0000 &     0.0000 &     0.0000 &     0.0000 &     0.0000 &     0.0000 &     0.0000 &     0.0000 &     0.0001 &     0.0001 &     0.0000 &            &            &            &            \\

MUSE &     0.0000 &     0.0000 &     0.0000 &     0.0000 &     0.0000 &     0.0000 &     0.0000 &     0.0000 &     0.0000 &     0.0000 &     0.0000 &     0.0000 &     0.0000 &     0.0000 &     0.0000 &     0.0000 &     0.0000 &     0.0329 &            &            &            \\

LEXRANK &     0.0000 &     0.0000 &     0.0000 &     0.0000 &     0.0000 &     0.0000 &     0.0000 &     0.0000 &     0.0000 &     0.0000 &     0.0000 &     0.0000 &     0.0000 &     0.0000 &     0.0000 &     0.0000 &     0.0000 &     0.0000 &     0.0000 &            &            \\

TEXTRANK &     0.0000 &     0.0000 &     0.0000 &     0.0000 &     0.0000 &     0.0000 &     0.0000 &     0.0000 &     0.0000 &     0.0000 &     0.0000 &     0.0000 &     0.0000 &     0.0000 &     0.0000 &     0.0000 &     0.0000 &     0.0000 &     0.0000 &     0.0423 &            \\

POLY &     0.0000 &     0.0000 &     0.0000 &     0.0000 &     0.0000 &     0.0000 &     0.0000 &     0.0000 &     0.0000 &     0.0000 &     0.0000 &     0.0000 &     0.0000 &     0.0000 &     0.0000 &     0.0000 &     0.0000 &     0.0000 &     0.0000 &     0.0003 &     \cellcolor{red}{0.1001} \\
\end{tabular}  
}
\caption{The p-values of the BERT score results using the Bootstrap test are presented in each column, where column i includes the p-values of system i and the p-values of the remaining n-i systems.}
\label{tab:BE-pvalues}

\end{table*}

\begin{table*}

\resizebox{\textwidth}{!}{%

\begin{tabular}{lccccccccccccccccccccc}

T5-Small-96 &            &            &            &            &            &            &            &            &            &            &            &            &            &            &            &            &            &            &            &            &            \\

LED-BASE-128 &     \cellcolor{red}{0.2622} &            &            &            &            &            &            &            &            &            &            &            &            &            &            &            &            &            &            &            &            \\

LED-BASE-256 &     \cellcolor{red}{0.1766} &     \cellcolor{red}{0.3579} &            &            &            &            &            &            &            &            &            &            &            &            &            &            &            &            &            &            &            \\

LED-BASE-1000 &     0.0774 &     \cellcolor{red}{0.2204} &     \cellcolor{red}{0.3232} &            &            &            &            &            &            &            &            &            &            &            &            &            &            &            &            &            &            \\

PEGASUS &     0.0000 &     0.0001 &     0.0004 &     0.0015 &            &            &            &            &            &            &            &            &            &            &            &            &            &            &            &            &            \\

BART &     0.0000 &     0.0000 &     0.0003 &     0.0011 &     \cellcolor{red}{0.4284} &            &            &            &            &            &            &            &            &            &            &            &            &            &            &            &            \\

mBART &     0.0000 &     0.0001 &     0.0001 &     0.0009 &     \cellcolor{red}{0.4227} &     \cellcolor{red}{0.4961} &            &            &            &            &            &            &            &            &            &            &            &            &            &            &            \\

LSA &     0.0000 &     0.0000 &     0.0000 &     0.0000 &     \cellcolor{red}{0.1042} &     \cellcolor{red}{0.1160} &     \cellcolor{red}{0.1261} &            &            &            &            &            &            &            &            &            &            &            &            &            &            \\

SBERT-EXTRACTIVE &     0.0000 &     0.0000 &     0.0000 &     0.0000 &     0.0010 &     0.0006 &     0.0011 &     0.0134 &            &            &            &            &            &            &            &            &            &            &            &            &            \\

T5-Small-256 &     0.0000 &     0.0000 &     0.0000 &     0.0000 &     0.0020 &     0.0013 &     0.0008 &     0.0142 &     \cellcolor{red}{0.4225} &            &            &            &            &            &            &            &            &            &            &            &            \\

T5-MULTI-REFERENCES &     0.0000 &     0.0000 &     0.0000 &     0.0000 &     0.0014 &     0.0008 &     0.0013 &     0.0113 &     \cellcolor{red}{0.3947} &     \cellcolor{red}{0.4955} &            &            &            &            &            &            &            &            &            &            &            \\

PEGASUS-MULTI-REFERENCES &     0.0000 &     0.0000 &     0.0000 &     0.0000 &     0.0008 &     0.0009 &     0.0015 &     0.0066 &     \cellcolor{red}{0.3410} &     \cellcolor{red}{0.4104} &     \cellcolor{red}{0.4304} &            &            &            &            &            &            &            &            &            &            \\

BERT-EXTRACTIVE &     0.0000 &     0.0000 &     0.0000 &     0.0000 &     0.0000 &     0.0000 &     0.0000 &     0.0004 &     \cellcolor{red}{0.1452} &     \cellcolor{red}{0.2396} &     \cellcolor{red}{0.2400} &     \cellcolor{red}{0.3016} &            &            &            &            &            &            &            &            &            \\

T5-Small-512 &     0.0000 &     0.0000 &     0.0000 &     0.0000 &     0.0000 &     0.0000 &     0.0000 &     0.0000 &     0.0233 &     0.0598 &     0.0539 &     0.0906 &     \cellcolor{red}{0.1589} &            &            &            &            &            &            &            &            \\

RNN-LSTM-RL &     0.0000 &     0.0000 &     0.0000 &     0.0000 &     0.0000 &     0.0000 &     0.0000 &     0.0000 &     0.0000 &     0.0004 &     0.0002 &     0.0005 &     0.0008 &     0.0177 &            &            &            &            &            &            &            \\

LEAD-1000 &     0.0000 &     0.0000 &     0.0000 &     0.0000 &     0.0000 &     0.0000 &     0.0000 &     0.0000 &     0.0000 &     0.0000 &     0.0000 &     0.0001 &     0.0000 &     0.0005 &     0.1594 &            &            &            &            &            &            \\

LED-LARGE &     0.0000 &     0.0000 &     0.0000 &     0.0000 &     0.0000 &     0.0000 &     0.0000 &     0.0000 &     0.0000 &     0.0000 &     0.0000 &     0.0000 &     0.0000 &     0.0007 &     0.1439 &     0.4598 &            &            &            &            &            \\

BART-MULTI-REFERENCES &     0.0000 &     0.0000 &     0.0000 &     0.0000 &     0.0000 &     0.0000 &     0.0000 &     0.0000 &     0.0000 &     0.0000 &     0.0000 &     0.0000 &     0.0000 &     0.0006 &     \cellcolor{red}{0.1427} &     \cellcolor{red}{0.4502} &     \cellcolor{red}{0.4757} &            &            &            &            \\

MUSE &     0.0000 &     0.0000 &     0.0000 &     0.0000 &     0.0000 &     0.0000 &     0.0000 &     0.0000 &     0.0000 &     0.0000 &     0.0000 &     0.0000 &     0.0000 &     0.0002 &     0.0586 &     0.3071 &     0.3319 &     0.3543 &            &            &            \\

LEXRANK &     0.0000 &     0.0000 &     0.0000 &     0.0000 &     0.0000 &     0.0000 &     0.0000 &     0.0000 &     0.0000 &     0.0000 &     0.0000 &     0.0000 &     0.0000 &     0.0000 &     0.0164 &     0.1609 &     0.1706 &     0.1829 &     0.2828 &            &            \\

POLY &     0.0000 &     0.0000 &     0.0000 &     0.0000 &     0.0000 &     0.0000 &     0.0000 &     0.0000 &     0.0000 &     0.0000 &     0.0000 &     0.0000 &     0.0000 &     0.0000 &     0.0015 &     0.0284 &     0.0433 &     0.0390 &     0.0723 &     0.1871 &            \\

TEXTRANK &     0.0000 &     0.0000 &     0.0000 &     0.0000 &     0.0000 &     0.0000 &     0.0000 &     0.0000 &     0.0000 &     0.0000 &     0.0000 &     0.0000 &     0.0000 &     0.0000 &     0.0000 &     0.0014 &     0.0004 &     0.0018 &     0.0026 &     0.0171 &     0.0946 \\

\end{tabular}  }

\caption{The p-values of the Bleurt score results using the Bootstrap test are presented in each column, where column i includes the p-values of system i and the p-values of the remaining n-i systems.}

\label{tab:Bleurt-pvalues}

\end{table*}

\begin{table*}

\resizebox{\textwidth}{!}{%

\begin{tabular}{lccccccccccccccccccccc}

T5-Small-96 &            &            &            &            &            &            &            &            &            &            &            &            &            &            &            &            &            &            &            &            &            \\

LED-BASE-128 &    \cellcolor{red}{0.1243} &            &            &            &            &            &            &            &            &            &            &            &            &            &            &            &            &            &            &            &            \\

LED-BASE-256 &     0.0777 &     \cellcolor{red}{0.1649} &            &            &            &            &            &            &            &            &            &            &            &            &            &            &            &            &            &            &            \\

LED-BASE-1000 &     0.0453 &     0.0874 &     \cellcolor{red}{0.2068} &            &            &            &            &            &            &            &            &            &            &            &            &            &            &            &            &            &            \\

   PEGASUS &     0.0000 &     0.0000 &     0.0000 &     0.0000 &            &            &            &            &            &            &            &            &            &            &            &            &            &            &            &            &            \\

T5-MULTI-REFERENCES &     0.0000 &     0.0000 &     0.0000 &     0.0000 &     0.0003 &            &            &            &            &            &            &            &            &            &            &            &            &            &            &            &            \\

T5-Small-256 &     0.0000 &     0.0000 &     0.0000 &     0.0000 &     0.0001 &     \cellcolor{red}{0.2906} &            &            &            &            &            &            &            &            &            &            &            &            &            &            &            \\

PEGASUS-MULTI-REFERENCES &     0.0000 &     0.0000 &     0.0000 &     0.0000 &     0.0000 &  \cellcolor{red}{0.1807} &     \cellcolor{red}{0.3979} &            &            &            &            &            &            &            &            &            &            &            &            &            &            \\

T5-Small-512 &     0.0000 &     0.0000 &     0.0000 &     0.0000 &     0.0000 &     0.0524 &     \cellcolor{red}{0.2062} &     \cellcolor{red}{0.2638} &            &            &            &            &            &            &            &            &            &            &            &            &            \\

      BART &     0.0000 &     0.0000 &     0.0000 &     0.0000 &     0.0000 &     0.0003 &     0.0023 &     0.0089 &     0.0187 &            &            &            &            &            &            &            &            &            &            &            &            \\

     mBART &     0.0000 &     0.0000 &     0.0000 &     0.0000 &     0.0000 &     0.0007 &     0.0021 &     0.0091 &     0.0191 &     0.0000 &            &            &            &            &            &            &            &            &            &            &            \\

BART-MULTI-REFERENCES &     0.0000 &     0.0000 &     0.0000 &     0.0000 &     0.0000 &     0.0000 &     0.0000 &     0.0000 &     0.0021 &     \cellcolor{red}{0.2190} &     \cellcolor{red}{0.2158} &            &            &            &            &            &            &            &            &            &            \\

 LED-LARGE &     0.0000 &     0.0000 &     0.0000 &     0.0000 &     0.0000 &     0.0000 &     0.0000 &     0.0000 &     0.0000 &     0.0526 &     0.0550 &     \cellcolor{red}{0.1431} &            &            &            &            &            &            &            &            &            \\

 LEAD-1000 &     0.0000 &     0.0000 &     0.0000 &     0.0000 &     0.0000 &     0.0000 &     0.0000 &     0.0000 &     0.0000 &     0.0273 &     0.0300 &     0.0724 &     0.4301 &            &            &            &            &            &            &            &            \\

RNN-LSTM-RL &     0.0000 &     0.0000 &     0.0000 &     0.0000 &     0.0000 &     0.0000 &     0.0000 &     0.0000 &     0.0000 &     0.0000 &     0.0000 &     0.0000 &     0.0000 &     0.0000 &            &            &            &            &            &            &            \\

      MUSE &     0.0000 &     0.0000 &     0.0000 &     0.0000 &     0.0000 &     0.0000 &     0.0000 &     0.0000 &     0.0000 &     0.0000 &     0.0000 &     0.0000 &     0.0000 &     0.0000 &     0.0000 &            &            &            &            &            &            \\

       LSA &     0.0000 &     0.0000 &     0.0000 &     0.0000 &     0.0000 &     0.0000 &     0.0000 &     0.0000 &     0.0000 &     0.0000 &     0.0000 &     0.0000 &     0.0000 &     0.0000 &     0.0000 &     0.0000 &            &            &            &            &            \\

SBERT-EXTRACTIVE &     0.0000 &     0.0000 &     0.0000 &     0.0000 &     0.0000 &     0.0000 &     0.0000 &     0.0000 &     0.0000 &     0.0000 &     0.0000 &     0.0000 &     0.0000 &     0.0000 &     0.0000 &     0.0000 &     0.3925 &            &            &            &            \\

BERT-EXTRACTIVE &     0.0000 &     0.0000 &     0.0000 &     0.0000 &     0.0000 &     0.0000 &     0.0000 &     0.0000 &     0.0000 &     0.0000 &     0.0000 &     0.0000 &     0.0000 &     0.0000 &     0.0000 &     0.0000 &     0.0070 &     0.0002 &            &            &            \\

   LEXRANK &     0.0000 &     0.0000 &     0.0000 &     0.0000 &     0.0000 &     0.0000 &     0.0000 &     0.0000 &     0.0000 &     0.0000 &     0.0000 &     0.0000 &     0.0000 &     0.0000 &     0.0000 &     0.0000 &     0.0000 &     0.0000 &     0.0000 &            &            \\

      POLY &     0.0000 &     0.0000 &     0.0000 &     0.0000 &     0.0000 &     0.0000 &     0.0000 &     0.0000 &     0.0000 &     0.0000 &     0.0000 &     0.0000 &     0.0000 &     0.0000 &     0.0000 &     0.0000 &     0.0000 &     0.0000 &     0.0000 &     0.0000 &            \\

  TEXTRANK &     0.0000 &     0.0000 &     0.0000 &     0.0000 &     0.0000 &     0.0000 &     0.0000 &     0.0000 &     0.0000 &     0.0000 &     0.0000 &     0.0000 &     0.0000 &     0.0000 &     0.0000 &     0.0000 &     0.0000 &     0.0000 &     0.0000 &     0.0000 &     0.0000 \\

\end{tabular}  }


\caption{The p-values of the Rouge-2 score results using the Bootstrap test are presented in each column, where column i includes the p-values of system i and the p-values of the remaining n-i systems.}
\label{tab:R2-pvalues}
\end{table*}

\begin{table*}




\resizebox{\textwidth}{!}{%

\begin{tabular}{lccccccccccccccccccccc}

T5-Small-96 &            &            &            &            &            &            &            &            &            &            &            &            &            &            &            &            &            &            &            &            &            \\

LED-BASE-256 &     \cellcolor{red}{0.1075} &            &            &            &            &            &            &            &            &            &            &            &            &            &            &            &            &            &            &            &            \\

LED-BASE-128 &     \cellcolor{red}{0.1330} &     \cellcolor{red}{0.7523} &            &            &            &            &            &            &            &            &            &            &            &            &            &            &            &            &            &            &            \\

LED-BASE-1000 &     \cellcolor{red}{0.0520} &     \cellcolor{red}{0.1245} &     \cellcolor{red}{0.0740} &            &            &            &            &            &            &            &            &            &            &            &            &            &            &            &            &            &            \\

   PEGASUS &     0.0000 &     0.0000 &     0.0000 &     0.0000 &            &            &            &            &            &            &            &            &            &            &            &            &            &            &            &            &            \\

T5-MULTI-REFERENCES &     0.0000 &     0.0000 &     0.0000 &     0.0000 &     0.0000 &            &            &            &            &            &            &            &            &            &            &            &            &            &            &            &            \\

PEGASUS-MULTI-REFERENCES &     0.0000 &     0.0000 &     0.0000 &     0.0000 &     0.0000 &     \cellcolor{red}{0.2039} &            &            &            &            &            &            &            &            &            &            &            &            &            &            &            \\

T5-Small-256 &     0.0000 &     0.0000 &     0.0000 &     0.0000 &     0.0000 &     \cellcolor{red}{0.2694} &     \cellcolor{red}{0.5332} &            &            &            &            &            &            &            &            &            &            &            &            &            &            \\

T5-Small-512 &     0.0000 &     0.0000 &     0.0000 &     0.0000 &     0.0000 &     \cellcolor{red}{0.1049} &    \cellcolor{red}{0.3750} &     \cellcolor{red}{0.3654} &            &            &            &            &            &            &            &            &            &            &            &            &            \\

      BART &     0.0000 &     0.0000 &     0.0000 &     0.0000 &     0.0000 &     0.0010 &     0.0138 &     0.0086 &     0.0159 &            &            &            &            &            &            &            &            &            &            &            &            \\

     mBART &     0.0000 &     0.0000 &     0.0000 &     0.0000 &     0.0000 &     0.0012 &     0.0156 &     0.0075 &     0.0157 &     0.0000 &            &            &            &            &            &            &            &            &            &            &            \\

BART-MULTI-REFERENCES &     0.0000 &     0.0000 &     0.0000 &     0.0000 &     0.0000 &     0.0000 &     0.0000 &     0.0008 &     0.0015 &     \cellcolor{red}{0.2342} &     \cellcolor{red}{0.2341} &            &            &            &            &            &            &            &            &            &            \\

 LEAD-1000 &     0.0000 &     0.0000 &     0.0000 &     0.0000 &     0.0000 &     0.0000 &     0.0000 &     0.0000 &     0.0000 &     0.0220 &     0.0204 &     0.0375 &            &            &            &            &            &            &            &            &            \\

 LED-LARGE &     0.0000 &     0.0000 &     0.0000 &     0.0000 &     0.0000 &     0.0000 &     0.0000 &     0.0000 &     0.0000 &     0.0483 &     0.0394 &     \cellcolor{red}{0.1052} &     \cellcolor{red}{0.5720} &            &            &            &            &            &            &            &            \\

RNN-LSTM-RL &     0.0000 &     0.0000 &     0.0000 &     0.0000 &     0.0000 &     0.0000 &     0.0000 &     0.0000 &     0.0000 &     0.0154 &     0.0156 &     0.0451 &     \cellcolor{red}{0.4762} &    \cellcolor{red}{0.4233} &            &            &            &            &            &            &            \\

      MUSE &     0.0000 &     0.0000 &     0.0000 &     0.0000 &     0.0000 &     0.0000 &     0.0000 &     0.0000 &     0.0000 &     0.0000 &     0.0000 &     0.0000 &     0.0000 &     0.0001 &     0.0000 &            &            &            &            &            &            \\

       LSA &     0.0000 &     0.0000 &     0.0000 &     0.0000 &     0.0000 &     0.0000 &     0.0000 &     0.0000 &     0.0000 &     0.0000 &     0.0000 &     0.0000 &     0.0000 &     0.0000 &     0.0000 &     0.0000 &            &            &            &            &            \\

SBERT-EXTRACTIVE &     0.0000 &     0.0000 &     0.0000 &     0.0000 &     0.0000 &     0.0000 &     0.0000 &     0.0000 &     0.0000 &     0.0000 &     0.0000 &     0.0000 &     0.0000 &     0.0000 &     0.0000 &     0.0000 &     0.0000 &            &            &            &            \\

BERT-EXTRACTIVE &     0.0000 &     0.0000 &     0.0000 &     0.0000 &     0.0000 &     0.0000 &     0.0000 &     0.0000 &     0.0000 &     0.0000 &     0.0000 &     0.0000 &     0.0000 &     0.0000 &     0.0000 &     0.0000 &     0.0000 &     0.0000 &            &            &            \\

   LEXRANK &     0.0000 &     0.0000 &     0.0000 &     0.0000 &     0.0000 &     0.0000 &     0.0000 &     0.0000 &     0.0000 &     0.0000 &     0.0000 &     0.0000 &     0.0000 &     0.0000 &     0.0000 &     0.0000 &     0.0000 &     0.0000 &     0.0031 &            &            \\

  TEXTRANK &     0.0000 &     0.0000 &     0.0000 &     0.0000 &     0.0000 &     0.0000 &     0.0000 &     0.0000 &     0.0000 &     0.0000 &     0.0000 &     0.0000 &     0.0000 &     0.0000 &     0.0000 &     0.0000 &     0.0000 &     0.0000 &     0.0000 &     0.0000 &            \\

  POLY &     0.0000 &     0.0000 &     0.0000 &     0.0000 &     0.0000 &     0.0000 &     0.0000 &     0.0000 &     0.0000 &     0.0000 &     0.0000 &     0.0000 &     0.0000 &     0.0000 &     0.0000 &     0.0000 &     0.0000 &     0.0000 &     0.0000 &     0.0000 &     0.0000 \\

\end{tabular}}


\caption{The p-values of the Rouge-L score results using the Bootstrap test are presented in each column, where column i includes the p-values of system i and the p-values of the remaining n-i systems.}
\label{tab:RL-pvalues}
\end{table*}

\end{document}